\def\year{2020}\relax
\documentclass[letterpaper]{article} 
\usepackage{aaai20nocopyright}  
\usepackage{times}  
\usepackage{helvet} 
\usepackage{courier}  
\usepackage[hyphens]{url}  
\usepackage{graphicx} 
\usepackage{graphicx}  
\usepackage{amsmath,amsfonts,subcaption}
\nocopyright
\title{TableNet: a multiplier-less implementation of neural networks for inferencing}
\author{Chai Wah Wu\\
IBM Research AI\\ IBM Thomas J. Watson Researcher Center\\ P. O. Box 218, Yorktown Heights, NY 10598, USA\\
cwwu@us.ibm.com
}

\begin{document}

\maketitle

\bibliographystyle{aaai}

\begin{abstract}
We consider the use of look-up tables (LUT) to simplify the hardware implementation of a deep learning network for inferencing after weights have been successfully trained. The use of LUT replaces the matrix multiply and add operations with a small number of LUTs and addition operations resulting in a completely multiplier-less implementation. We compare the different tradeoffs of this approach in terms of accuracy versus LUT size and the number of operations and show that similar performance can be obtained with a comparable memory footprint as a full precision deep neural network, but without the use of any multipliers. We illustrate this with several architectures such as MLP and CNN.
\end{abstract}

\section{Introduction}
In recent years, neural networks (NN) \cite{LeCun2015} have re-emerged as a powerful tool in many nonlinear regression and classification applications, especially in the visual \cite{Voulodimos2018} and audio \cite{Gavat2015} processing domains. The general form of a neural network can be expressed as a sequence of linear and nonlinear mappings. In particular, we consider the following $N$-layer feed-forward network formulation:

\begin{equation} y_{i+1} = f_i(W_iy_i + b_i), i = 1,..., N
\label{eqn:nn}
\end{equation}
where $f_i$ is a nonlinear function mapping vectors of length $m_i$ to vectors of lengths $n_{i+1}$. They can be vectorized activation functions (ReLu, sigmoid, tanh, etc.), but can also be other functions such as pooling, softmax and even linear functions such as the identity function. Each $W_i$ is a matrix of length $m_i\times n_i$ and $y_i$ is a vector of length $n_i$, although for easier visualization and interpretation, they can be represented as (or reshaped into) a multidimensional tensor with the same number of elements. The vectors $b_i$ are of length $m_i$.
The vectors $y_1$ and $y_N$ are the input and the output vectors of the entire network respectively and thus the network in Eq. (\ref{eqn:nn}) describes a nonlinear mapping between input and output. The functions $f_i$ are assumed to be fixed and given. The goal in training a network for classification and regression is that given $K$ pairs of vectors $(x_k, z_k)$, find the set of weights $W_i$ and biases $b_i$ such that $\sum_k d(\tilde{z}_k, z_k)$ is minimized, where $\tilde{z}_k$ is the output from the neural network when $x_k$ is fed as input to the neural network and $d(\cdot,\cdot)$ is an error or loss metric. 

After the network is trained, Eq. (\ref{eqn:nn}) is used in the inference phase to compute $y_{N+1}$ given an new input data element $y_0$. Since this may be done in a low-power environment or at high speed, there is a need to speed up the computation of Eq. (\ref{eqn:nn}) using low-power high speed hardware. The evaluation of Eq. (\ref{eqn:nn}) consists of many multiply-and-add operations and nonlinear operations such as sigmods and logistic equations which also require many arithmetic operations. Modern GPU architectures have been adopted to perform these computations with a large number of arithmetic units and digital multipliers. The purpose of this paper is to use look-up tables (LUT) to construct a truly multiplierless implementation that is easily parallelizable. While this is the primary motivation of this work, we show that we can achieve comparable accuracy with a similar memory footprint as a reference neural network implementation.

\section{LUT framework and notation}
A LUT is a memory array to store precomputed values of a typically complicated function and has been found useful in high performance image processing to map between different color spaces where this mapping is generally nonlinear. In particular, current GPUs support the use of 2.5D and 3D LUT for color conversion at high speed \cite{pharr:gpugems2}. 

We can consider a LUT as a function $f:I\rightarrow O$ that maps elements from an input set $I$ to elements of an output set $O$. 
The number of bits required to describe a element $x\in I$ is denoted $\beta(I)$ (we will also denote this as the {\em resolution} of $I$) and is given by $\beta(I) = \lceil \log_2 (|I|)\rceil$ where $|I|$ is the cardinality of $I$. We assume that the bits used to describe elements of $I$ are in a number format so that arithmetic operations can be performat directly on the bits.
For instance if $I$ are integers from $0$ to $255$, then $\beta(I) = 8$. If $I$ are IEEE 754 single precision floats (as used in the C programming language), then $\beta(I) = 32$. If $I$ are 8-bit minifloats \cite{wikipedia:minifloat}, then $\beta(I) = 8$. 
Thus the LUT is indexed by $\beta(I)$ bits and output $\beta(O)$ bits and the size of a LUT is $2^{\beta(I)}\beta(O)$ bits.

\subsection{Partitioning the input bits}
 Note that there is an asymmetry between $I$ and $O$ in determining the size of the LUT. This suggests that $\beta(I)$ should be as small as possible and typically $\beta(O)$ is larger than $\beta(I)$.
If $\beta(I)$ is too large, one way to reduce the resulting LUT is to partition the bits in $\beta(I)$. We assume that the bits in an element of $I$ are additive in the sense that if $b_i$ are the bits in $x\in I$, then $x = \sum_i b_i\alpha_i$ for some fixed objects $\alpha_i$. This is certainly the case when $x$ is a vector, matrix or tensor of numbers in fixed point or floating point formats, in which case $\alpha_i$ are also vectors, matrices or tensors. 
If we partition the bits into $k$ chunks of  $m_i$ bits such that $\sum_{i=1}^k m_i = \beta(I)$, then we can construct $k$ LUTs of
size $2^{m_i}\beta(O)$ bits each. The bits in $x\in I$ are partitioned into $k$ chunks, applied to these LUTs and their results added (according to $x = \sum_i b_i\alpha_i$). Note that generally $m_i \geq 2$; if we split a 2-bit chunk into 2 1-bit chunks, we did not reduce the total LUT size as  $2^{2}\beta(O) = 2^{1}\beta(O) + 2^{1}\beta(O)$.

From the previous section, a NN can be decomposed into 2 types of modules, the affine operation $Wx+b$ and the nonlinear function $f$. Each of these modules can be replaced with LUTs. We describe each of these in turn.

\section{Computing  a nonlinear function $f$ with LUT}
Replacing a general nonlinear function $f:I\rightarrow O$ with a LUT is generally feasible only if $\beta(I)$ is small.
This means that using a single LUT is more suitable for scalar functions $f: \mathbb{R} \rightarrow \mathbb{R}$ such as sigmoids and activation functions or pooling layers in later stages of a NN where the information is more compressed into features. For instance, a scalar function that maps $32$-bit floats to $32$-bit floats can be implemented with a LUT table of size $2^{37}$ bits or $16$ Gibibytes\footnote{We use here the ISO/IEC 80000 International System of Quantities units \cite{vim3:2012} for quantifying the size of digital information: kibibytes (KiB) = $2^{10}$ bytes, mebibytes (MiB) = $2^{20}$ bytes, gibibytes (GiB) = $2^{30}$ bytes.} which is quite unwieldy. However, reducing the input and output to a 16-bit half-precision float reduces the LUT table size to $128$ Kibibytes. It is possible to search for a nonlinear way to combine the output of multiple LUTs to approximate $f$ in order to reduce further the total size of the LUT, but that is not the focus of this paper as it is a much more difficult problem. 
In many recent NN architectures, the activation function is a rectified linear function (ReLu) which can simply be implemented with a compare and branch (either in software or hardware) instruction and does not need the use of a LUT.

\section{Computing the affine operation $Wx + b$ and exploiting linearity}
The most computation-intensive part of a NN is the affine operation $Wx+b$.
We can exploit linearity to compute $Wx+b$ efficiently using LUTs.
Let $W$ be a $p$ by $q$ matrix, $b$ a $p$-vector. Let $x\in I$  be a $q$-vector where each element is represented with $r_I$ bits. Thus $\beta(I) = qr_I$. The output $Wx+b$ is a $p$-vector whose elements are represented with $r_O$ bits each, i.e. $\beta(O)=pr_O$.
We first partition the vector $x$ into $k$ segments $x_i$ of size $m_i$ such that $\sum_{i=1}^k m_i = q$.\footnote{To simplify the presentation, there is a bit of abuse of notation here. What we mean here is that $x_i$
 is obtained from $x$ by setting some elements of $x$ to be $0$, i.e. $x_i$ is the same length as $x$, but some of the entries are zero.
 The elements of $x$ are partitioned in such a way such that the number of elements of $x_i$ that were set to $0$ is $m_i$ and
$x =\sum_i x_i$. In this way $x_i$ can also be interpreted as a segment of $x$ if we only looked at its nonzero elements and vector
addition corresponds to concatenation of the vector segments.}
Then each of these segments $x_i$ is used to build a LUT which outputs $Wx_i + \frac{1}{k} b$. 
The output of these table lookup operations is then added to obtain the final result $Wx+b = \sum_{i=1}^k Wx_i + b$.
The total size of the LUT tables is $\sum_{i=1}^k 2^{m_ir_I}r_O$ bits and a total of $k-1$ additions of $p$-vectors is needed.
This is in contrast with $pq$ multiply and add operations for a standard implementation of $Wx+b$. In particular, if we choose $k = q$, $m_i = 1$, we will have $q$ LUT tables with a total size of $2^{r_I}qr_O$ bits and $q-1$ additions of $p$-vectors. Thus the number of additions is the same as the standard implementation, but all the $pq$ $r_I$-bit multiplications are replaced with $q$ LUT operations.

\subsection{Fixed point formats} \label{sec:fixedpoint}
If $x$ is stored in a fixed point format, additional simplification and efficiency can be obtained by exploiting linearity in the fixed point representation.
Consider a  $q$-vector $x$ where each element is denoted $x_i$. Let $r_I = n$, i.e. each number $x_i$ is represented as a $n$-bit number $x_i = \sum_{j=0}^{n-1} a_{ij} 2^j$ where $a_{ij}$ are the bits representing $x_{ij}$.
The linear combination $y = \sum_{i} w_i x_i $ can then be written as  $y =\sum_{i} \sum_{j=0}^{n-1} w_i a_{ij} 2^j$. Swapping the order of summation we get $\sum_{j=0}^{n-1} 2^j \sum_{i} w_i a_{ij}$. For a fixed $j$, the bits $a_{ij}$ correspond the $j$-th bitplane of the numbers $x_j$. This implies that we can use the {\em same} LUT for each bitplane and do $n$ shift-and-add operations to compute $y$. 

Again $q$ is partitioned into $k$ segments of size $m_i$ such that $\sum_{i=1}^k m_i = q$.
Then each of the $m_i$ elements of $x$ is used to construct a LUT which output $Wx_{ij} + \frac{1}{k} b_j$
Thus the total numbers of bits for the LUT is $\sum_{i}2^{m_i}b(O)$ and $nk$ LUT evaluations and shift-and-add operators are needed to compute $Wx+b$. In this scenario, as mentioned before, $m_i\geq 2$ as there is no advantage in choosing $m_i = 1$.

Instead of using a single bitplane in the input of a LUT, a subset of bitplanes can be used to index the LUT. In this case, the set of bitplanes is partitioned into blocks that can be shifted into each other  (e.g. pairs of adjacent bits).

\subsection{Floating point formats} \label{sec:bfloat}
\begin{figure}[htbp]
\centerline{\includegraphics[width=0.4\textwidth]{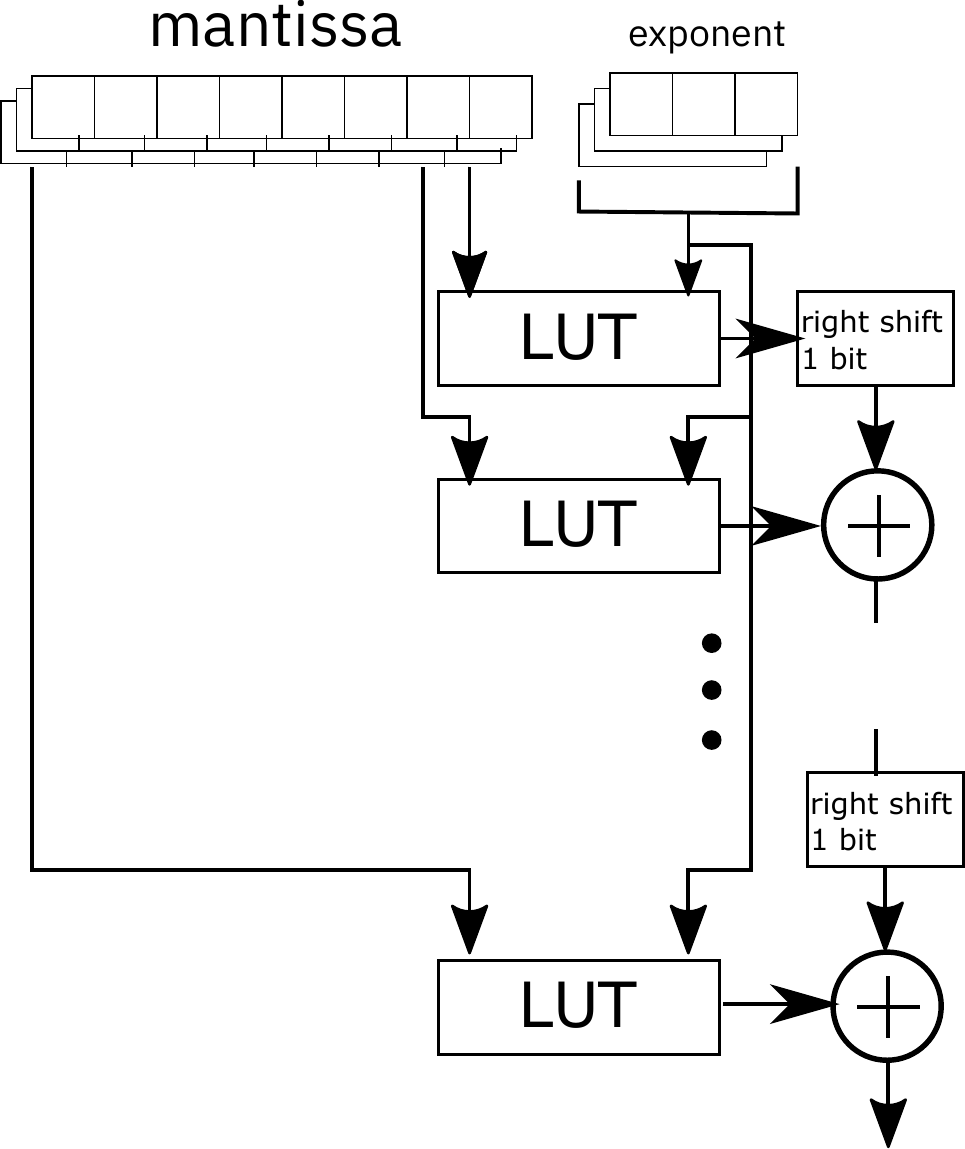}}
\caption{Splitting the input when the data is represented in floating point format. The mantissa is split into bitplanes whereas the entire exponent is input into the LUT. The same LUT is used for each bit of the mantissa.}\label{fig:fp}
\end{figure}

When numbers are represented in floating point format, similar to the fixed point format, we can also split the mantissa into bitplanes (or groups of bitplanes). However, the entire exponent need to be part of the input to each LUT. For instance, consider a floating point representation with $r_I$ bits where $r_I = n + t$, with $n$ bits reserved for the mantissa and $t$ bits reserved for the exponent. The same LUT is used to index a single bitplane of the mantissa and the entire $t$ bits of the exponent. This is illustrated in Fig. \ref{fig:fp}. With $k$ and $m_i$ defined as above, the total number of bits for the LUTs is $\sum_{i}2^{m_i(1+t)}b(O)$ and $nk$ LUT evaluations and bit-shift-and-add are needed to compute $Wx+b$. This suggests that in order to obtain a small total LUT size, the number of bits allocated to the exponent should be small.

\subsection{Convolutional layers using LUT}

\begin{figure}[htbp]
\centering
\centerline{\includegraphics[width=0.45\textwidth]{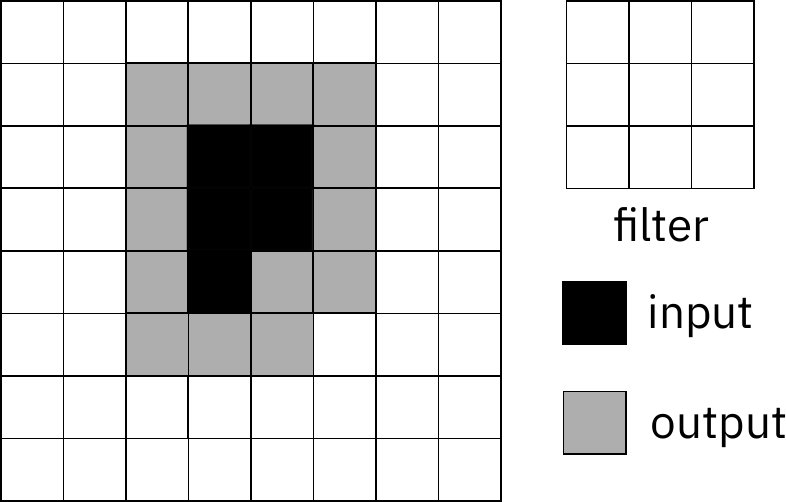}}
\caption{Filter size, support of input, and the support of output.}\label{fig:convlayer}
\end{figure}

In convolutional layers the $W$ matrix has a specific structured form and many of the matrix coefficients are duplicated. For instance, a 1-D convolutional layer has a circulant matrix $W$ whereas in 2-D convolutional layers, the matrix $W$ is a block-circulant matrix of circulant blocks. To implement this in a LUT, first the input $q$-vector is partitioned into $k$ chunks whose support are shifted version of each other. This could be either contiguous chunks or alternating elements. The output will be a vector that is larger related to the filter size (Fig. \ref{fig:convlayer}) and is equal to the dilation of the input support with the convolutional filter structural element \cite{Gonzalez1992}. To minimize the size of the output support it  is better to have the partition be in square contiguous blocks. The shift-invariance in the linear convolution operation is analogous to the invariance of the linear operations on the bitplanes and we can similarly reuse the same LUT. 
In particular, the same LUT can be used for each of the chunks and the output shifted (in space here rather than in binary base (Sect. \ref{sec:fixedpoint})) and added.
Suppose that due to the convolution an $a$-element vector is mapped to an output vector of size $c$. Then the LUT size is
$2^{ar_I}cr_O$ bits. For example, in a typical $2r+1$ by $2r+1$ 2-D convolutional filter, for an input block of $m$ by $m$ pixels with $a=m^2$, the output block will be $m+2r$ by $m+2r$ pixels with $c = (m+2r)^2$.

\begin{figure}[htbp]
\centerline{\includegraphics[width=0.45\textwidth]{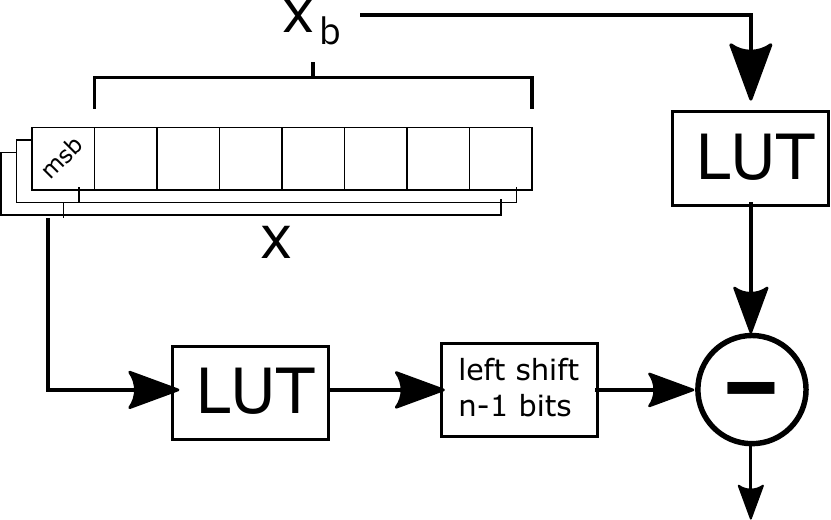}}
\caption{Processing signed numbers. The same LUT is used for both $x_b$ and the MSB.}\label{fig:signed}
\end{figure}

\subsection{Dealing with signed numbers}
So far the discussion deals only with using unsigned numbers in the input set $I$ to index the LUT\footnote{The output set $O$ can be represented in any number format (signed or unsigned). However since the output of a layer in a NN is the input to the next layer, the input needs to accomodate both signed and unsigned numbers as well.}. Dealing with signed numbers requires a slight modification of the architecture. We discuss here the case of fixed point formats as the floating point format is similarly handled. Consider a $n$-bit bitstring $x$ encoding a number in 2's complement format. The most significant bit (MSB) is a sign bit. If this bit is 1, then the value represented by $x$ is $x-2^n$ = 
$x_b - 2^{n-1}$ where $x_b$ is the bitstring $x$ minus the MSB. Depending on whether the MSB is 1 or not, there is an additional offset of $-2^{n-1}$. Thus we can partition the set of $x_b$'s as before, apply the LUTs (this can be applied to the entire bitstring $x_b$ or by bitplanes as described above), and add the results. The MSB of all the elements in the vectors are similarly partitioned and applied to the same LUTs, and the result shifted to the left by $n-1$ bits and subtracted from the previous result. This is shown schematically in Fig. \ref{fig:signed}.

In many NN architectures, there is no need to deal with signed numbers, since the input to a linear layer is generally the output of an activation layer or a pooling layer. Many activation functions are either nonnegative (e.g. ReLu) or can be made nonnegative by the addition of a fixed positive constant. This means that the pooling layer after an activation layer will also be nonegative.

\section{Stochastic rounding}
A LUT can also be used to implement stochastic rounding that has been found useful in ML algorithms using limited precision \cite{Gupta2015}. The rounding function is augmented with an additional input as a counter. Let $r(i)$ be a sequence of $R$ (pseudo)random numbers between 0 and 1\footnote{$r(i)$ can also be chosen using a 1-d dithering or halftoning algorithm \cite{ulichney_halftoning:1987}.}, then the function that is  implemented by the LUT is:
 \[
f(x,i) = \left\{\begin{array}{ll} \lfloor x \rfloor & \mbox{if }  r(i) \leq 1+ \frac{\lfloor x\rfloor - x}{\epsilon} \\ 
					 \lfloor x \rfloor+\epsilon & \mbox{otherwise}
					 \end{array}
					 \right.
\]
The index $i$ is incremented (modulo $R$) each time the LUT table is accessed. The size of the LUT is $R2^{\beta(I)}\beta(O)$ bits.
\section{Example implementations}
We consider multiple neural network architectures for the tasks of classifying the MNIST dataset \cite{LeCun2010} and the Fashion MNIST dataset \cite{xiao_fashion-mnist:_2017}.
We insert quantization operations before the input to a CNN or dense linear layer to mimic the quantization required to map it to the desired input set $I$ for the input of a LUT. We trained this modified network using Tensorflow and SGD with dropout. The ReLu activation layers, the pooling layers, and the argmax layer to determine the label from the one-hot encoding do not involve any multiplication and only use comparison operations only. We will omit these layers from our comparison since they are the same for the LUT approach and the traditional approach.
Because of the ReLu activation layers, the sign bit in the input $x\in I$ to the LUT will always be $0$ so can reduce the LUT size by half when using a floating point format as the input.

\subsection{Linear classifier}
Consider a linear classifier with a single dense layer $(W,b)$ of sizes $784\times 10$ and $10\times 1$ respectively. The total storage for the weight matrices in single precision floating point format is $30.7$Kibibytes. We run the training for 50000 episodes with a minibatch size of 100 and averaged the results over 20 trials.
\subsubsection{MNIST}
\begin{figure}[htbp]
\centerline{\includegraphics[width=0.5\textwidth]{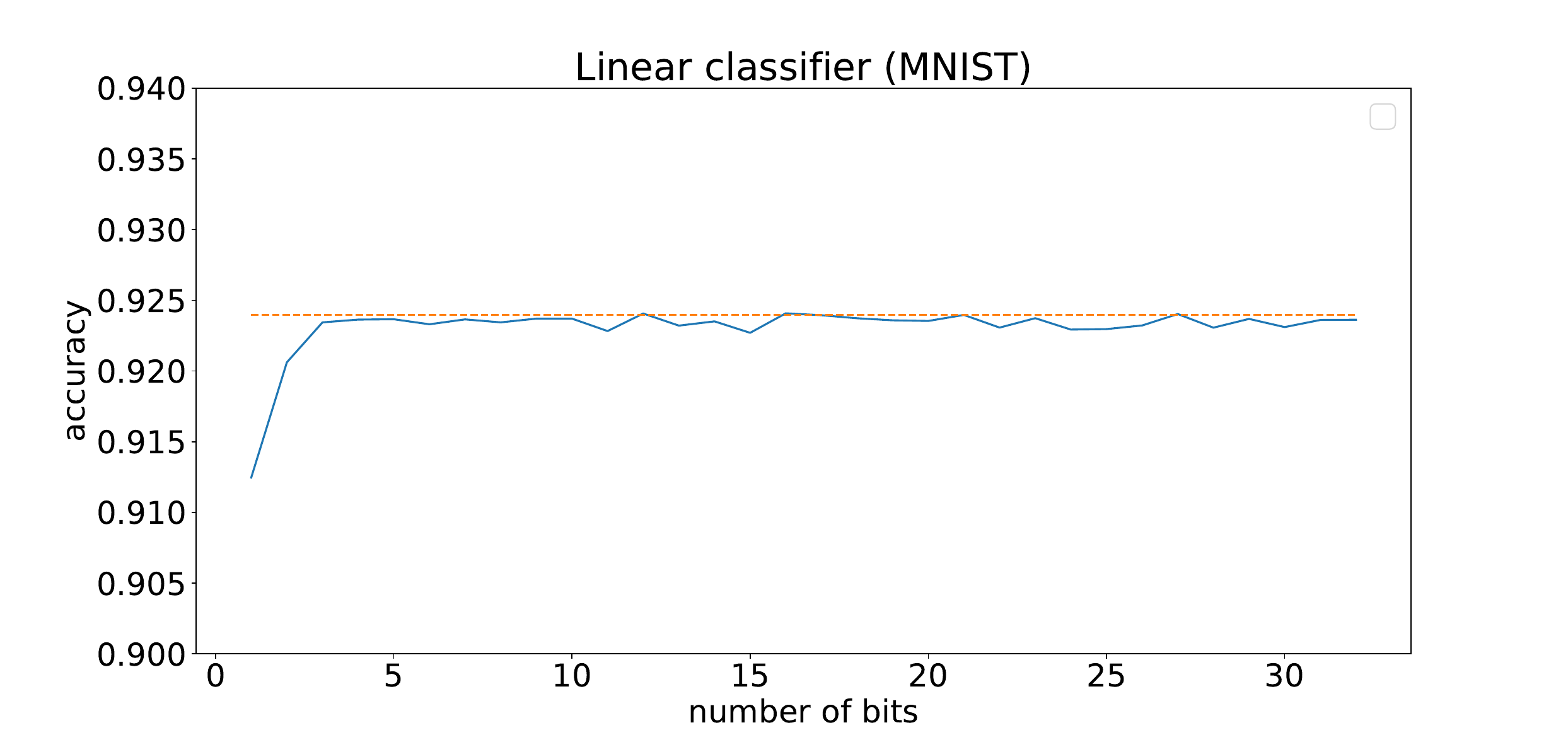}}
\caption{Number of bits in input versus accuracy on MNIST data using a linear classifier. The orange horizontal line shows the accuracy of the reference full precision linear classifier.}\label{fig:mnist-linear}
\end{figure}

The reference model achieves an average accuracy of $92.4\%$ on the test data set.
For the LUT-based implementation, the LUT implements the operations $Wx+b$ and accepts $x$ as an input in fixed point format.
The accuracy versus the number of bits in the input is shown in Fig. \ref{fig:mnist-linear}.

\begin{figure}[htbp]
\centerline{\includegraphics[width=0.5\textwidth,clip=true,trim=200 72 200 72]{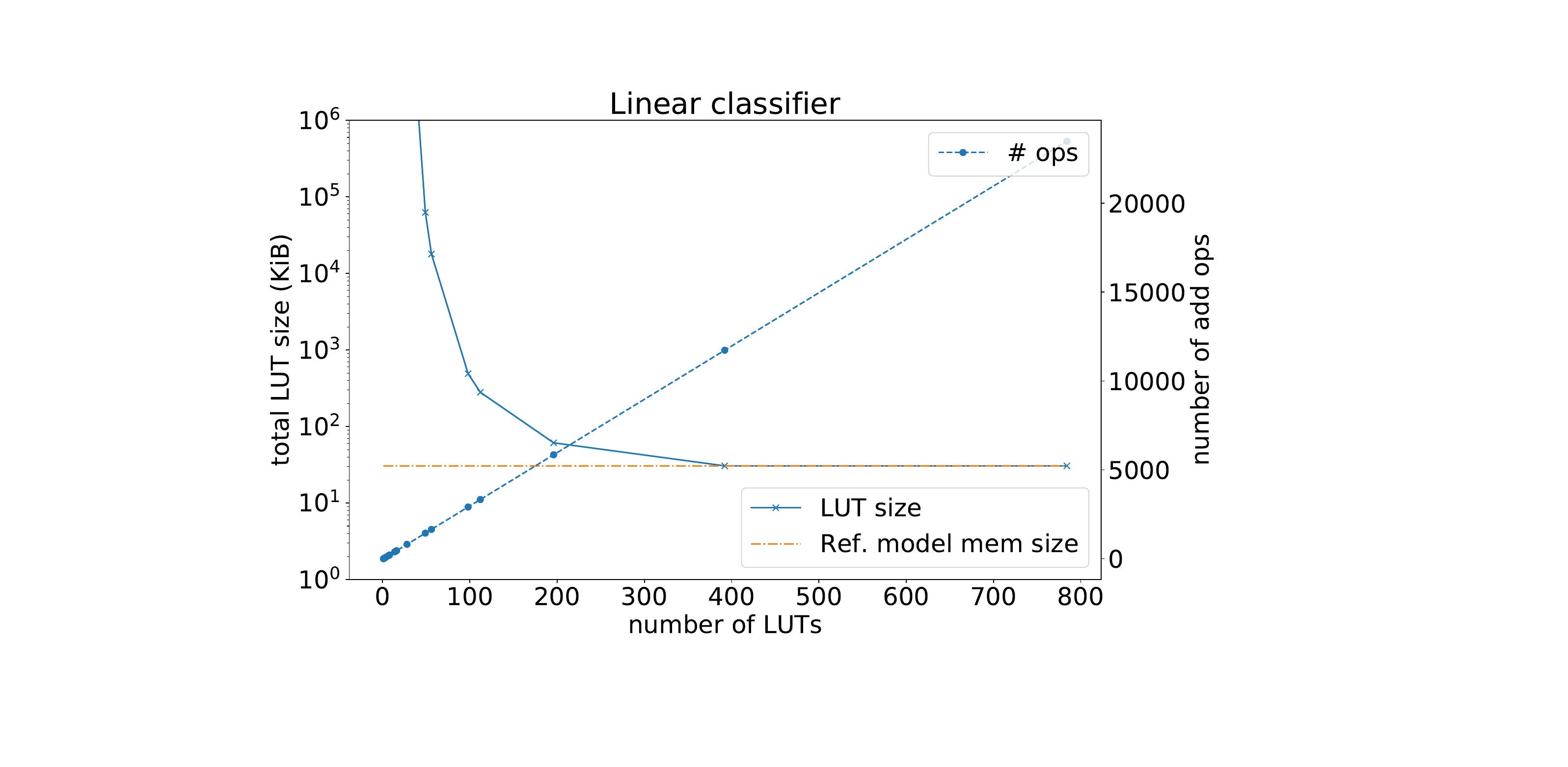}}
\caption{Tradeoff between total LUT size versus number of shift-and-add operations for inference on MNIST and Fashion MNIST data using a linear classifier.}\label{fig:slpops}
\end{figure}

With the input quantized to about 3 bits, we were able to achieve similar accuracy and that increasing the precision on the input does not increase the accuracy noticeably. This is not surprising since the original NIST digits images \cite{Grother1995} are bilevel and the few grey levels were introduced into MNIST due to anti-aliasing. Quantizing the input to 3 bits implies that the totality of bits in the input of the LUT is $3\times 28\times 28=2352$. The output is a one-hot encoding of the classified label ranging from 1 to 10, i.e. it can be represented with 10 16-bit half-precision float numbers.
Partitioning these input bits into various partition result in the tradeoff of LUT size versus number of shift-and-add operations (Fig. \ref{fig:slpops}). For instance, we can perform each inference on MNIST using 56 LUTs with a total combined size of 17.5 Mebibytes, 168 LUT evaluations and 1650 shift-and-add operations compared to 7840 multiply and add operations in the referencel model. In fact, using $784$ LUTs totaling about $30.6$Kibibytes, the number of shift-and-add operations is $23520$ and has the same memory footprint as the reference model but without any multiplications involved.

\subsubsection{Fashion MNIST}
\begin{figure}[htbp]
\centerline{\includegraphics[width=0.5\textwidth]{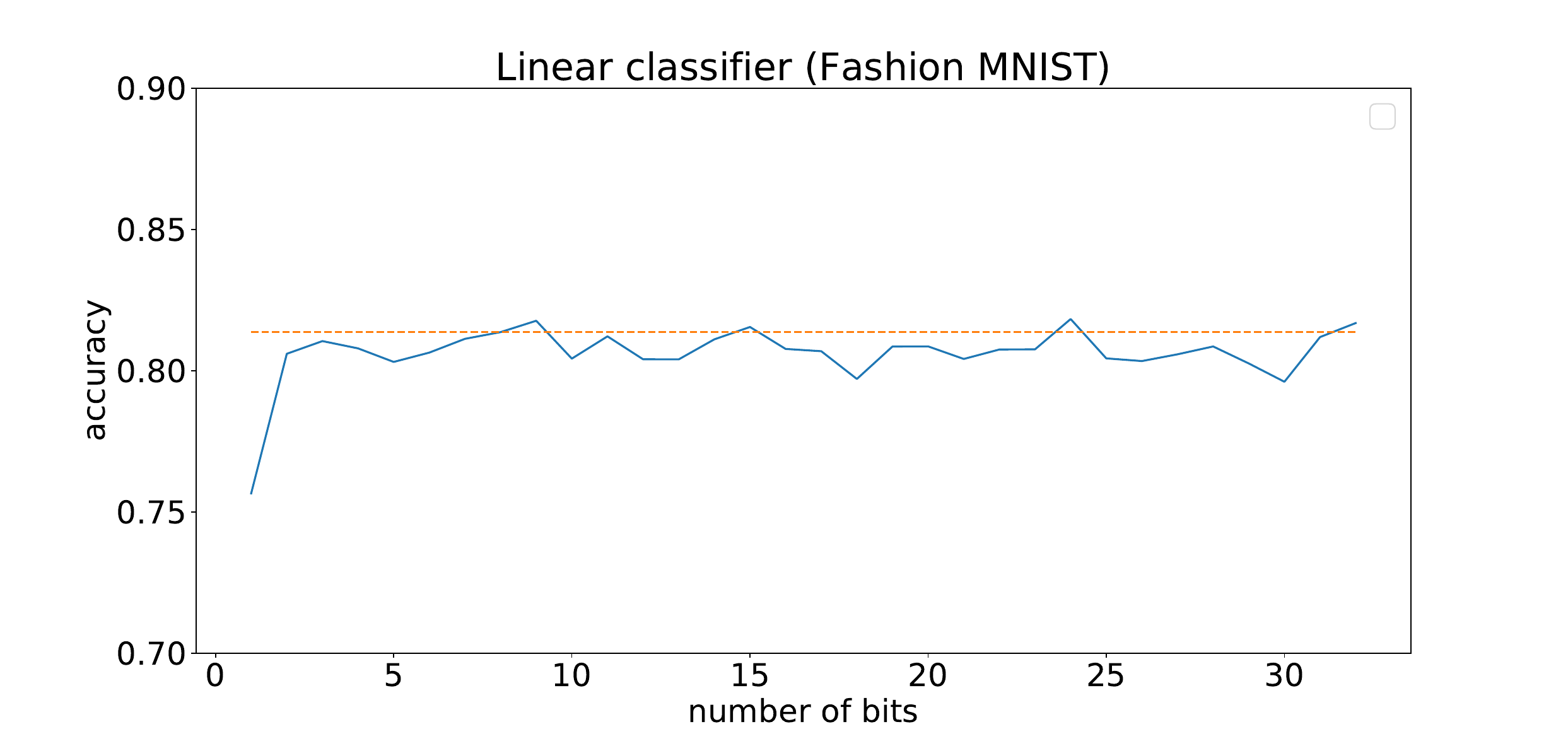}}
\caption{Number of bits in input versus accuracy on Fashion MNIST data using a linear classifier. The orange horizontal line shows the accuracy of the reference full precision linear classifier.}\label{fig:fmnist-linear}
\end{figure}

After training, the reference model using 32-bit single precision floating point arithmetic achieves an average accuracy of $81.4\%$ on the test dataset.
The tradeoff in accuracy versus the number of bits in the input is shown in Figure \ref{fig:fmnist-linear}.
Similarly, we see that quantizing the input to 3 bits per pixel suffices to reach similar accuracy. 
Interestingly, we see that the accuracy can decrease slightly as the number of bits increase. 
This we believe is due to the fact that the loss of information in quantization counteracts the decrease in accuracy on the testing tasks due to overfitting on the training tasks. 
Similar to the MNIST case we can perform each inference on Fashion MNIST using 56 LUTs with a total combined size of 17.5 Mebibytes, 168 LUT evaluations and 1650 shift-and-add operations.

\subsection{Multilayer Perceptron (MLP)}
We consider a 3-layer neural networks with 3 dense layers of sizes ($784\times 1024$, $1024\times 1$), ($1024\times 512$,  $512\times 1$) and ($512\times 10$,  $10\times 1$) respectively. These weights requires about 5.1 Mebibytes in storage.
For the MNIST dataset, the reference model achieves 98.2\% accuracy.
We will use a 8-bit fixed point format to encode the input image pixels for the first dense layer.
Using a fixed point format for the input to the second and third dense layers result in a reduced accuracy.

On the other hand, using IEEE 754 binary16 16-bit floating point format for the output of the first layer and the second layer and the input to the second and third layer, we obtain an accuracy of 98.4\% which is comparable to the reference model.

There are two ways to implement the float point format architecture. If all 16 bits are used to index the LUT, 
we can achieve similar performance as the reference model, with 2320 LUTs with a combined size of 32.7 Gibibytes and 1330678 addition operations compared with 1332224 multiply-and-add operations. This LUT size is not practical in current implementations. 

\begin{figure}[htbp]
\centerline{\includegraphics[width=0.5\textwidth,clip=true,trim=200 72 200 72]{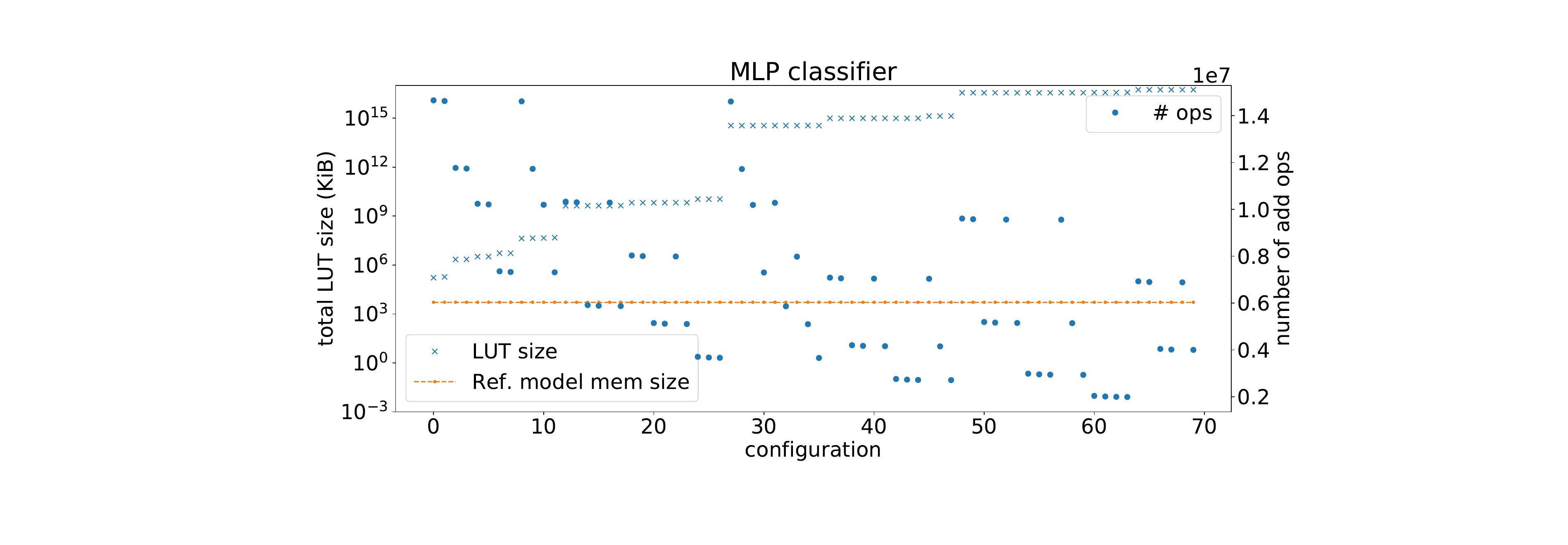}}
\caption{Tradeoff between total LUT size versus number of addition operations for inference on MNIST data using a MLP classifier. The configurations are sorted according to total LUT size.}\label{fig:mlpops}
\end{figure}

The second way is to split the mantissa into bitplanes and apply the same LUT to each bitplane and apply a shift-and-add as described in Sect. \ref{sec:bfloat}.
The precision in the mantissa of the IEEE 754 binary16 format is 11 bits. The sign bit is always 0 since we are only dealing with nonnegative numbers due to the use of ReLu activation.
If the mantissa are separated into these 11 bit-planes but still use the entire 5-bit exponent to index the LUT, the tradeoffs are shown in Fig. \ref{fig:mlpops}.
Thus we can achieve similar performance as the reference model, with 2320 LUTs with a combined size of 162.6 Mebibytes and 14652918 shift-and-add operations compared to 1332224 multiply-and-add operations in the reference model.

For the Fashion MNIST dataset, the reference model achieves 89.7\% accuracy and similar to MNIST, using binary16 in the input we obtained a similar accuracy.

\subsection{Deep CNN}
Consider next a LeNet implementation for classifying MNIST that was described in a Tensorflow tutorial. The weight matrices for the dense and convolutional linear layers (with pooling and dropout layers in between) are:
\begin{enumerate}
\item Convolutional layer 1 with filter size 5x5 and outputing 32 features
\item Convolutional layer 2 with filter size 5x5 and outputing 64 features
\item Dense layer 1 of size 3136x1024
\item Dense layer 2 of size 1024x10
\end{enumerate}
This reference model using single precision arithmetic achieves an accuracy of about 99.2\%. The weight matrices take up about 12.49 Mebibytes. The number of multiply-and-add operations are 12.9M.

\begin{figure}[htbp]
\centerline{\includegraphics[width=0.45\textwidth,clip=true,trim=200 72 200 72]{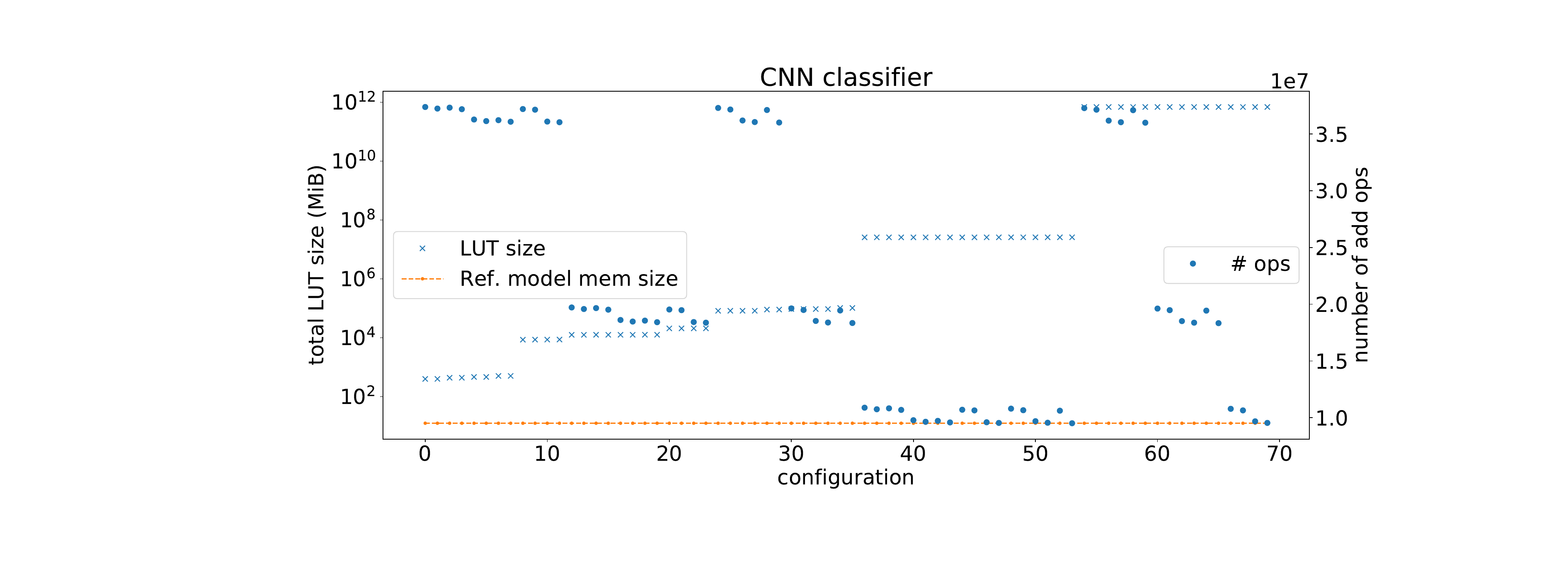}}
\caption{Tradeoff between total LUT size versus number of shift-and-add operations for inference on MNIST data using a CNN classifier. The configurations are sorted according to total LUT size.}\label{fig:cnnops}
\end{figure}

Again, we use 8-bits in fixed point format to encode the input images to the network. We find that with a fixed point format indexing the input to layers 2 through 4, we can only get an accuracy of about 95.6\%. On the other hand using the IEEE 354 binary16 floating point format to encode the input to layers 2 through 4, we achieved similar accuracy as the reference model.
The smallest total LUT size is achieved for the configuration where the mantissa is partitioned into 11 bitplanes and the spatial partition is into single elements. In this case, the total LUT size is 400Mebibytes. The number of shift\footnote{Note that the shift here are of two kinds: spatial (due to convolutions) and along the bits in binary (due to the bitplanes).} and add operations are 37.4M.
A more detailed tradeoff of total LUT size versus number of operations is shown in Fig. \ref{fig:cnnops}. For instance, another partitioning configuration results in a total LUT size of 12.26Gibibytes and 12.9M shift-and-add operations (comparable to the number of multiply-and-add operations in the reference model).

Even though we focus on MNIST for our experimental results, this approach scales
proportionally to other Deep Learning networks, i.e. if a network with limited resolution input can solve the problem
with similar accuracy, then the corresponding LUT approach will solve it just as well.

\section{Comparison with low precision NN}
While the philosophy of trading off accuracy versus performance by leveraging low precision data in the present approach is at first glance similar to approaches using low-precision data format and arithmetic (e.g. \cite{Gupta2015} or tensorflow-lite) or binary weights and activations \cite{Hubara2016}, there are some significant and fundamental differences between the two approaches. First of all, because of the asymmetry between the input and output in a LUT, the main reduction in the precision is the input $I$ in a LUT. The weights, the nonlinearity, the output $O$ and the computations needed to produce the elements in $O$ that are stored in the LUT can all be done in full precision. This is in contrast to limited precision approaches   where the entire computation pipeline (including the storage and representation of the weights) is done in low precision. Secondly, because the LUT is indexed by the totality of bits in $I$, where $I$ can be an ordered set of multiple elements, the precision of different elements in $I$ can be different. This is in contrast to using low precision arithmetic hardware where all the computation is done at the same precision. For instance, Ref. \cite{Wu2006} shows an application in image halftoning using LUT where the precision of the input depends on the location of the input pixel.
In terms of parallism, the LUT approach can be parallelized as the LUT operations can be done in parallel with multiple LUTs, whereas the paralellism in multiplier-based NN replies on using a parallel multiplier (rather than an iterative shift and add multiplier) and having several multiplier hardware running in parallel.

\section{Concluding remarks} 
We explore the feasibility of using LUT to speed up the inference of  large scale neural networks by eliminating multiply and add operations, especially in CNNs where the majority of computations during inference is computing the convolutions.
This approach can be implemented in software using standard computing hardware. However, custom hardware for partitioning the input data  (whether in floating point or fixed point format) into chunks that can be used directly as indices to the LUT and for bit-shuffling operations applied to both the input and the output of the LUT can result in an even more efficient architecture. Such custom hardware would simply reroute the bits appropriately to access memory locations of the LUT and rerouting the output from the LUT appropriately to the adder.
We believe that in future mobile Internet-of-Things (IoT) or edge computing environments, where data is acquired at the sensors at a very high rate, it makes sense to have computation done at the sensor level. In these scenarios having a LUT at each sensor may be an effective solution.

Our experimental results indicate that a practical tradeoff between total size of LUT, number of arithmetic operations and accuracy exists.
In general, having a large total LUT size reduces the number of operations, and we show that it is possible to 
have a multiplier-less LUT implementation with a similar memory footprint as a traditional solution requiring multipliers while achieving similar accuracy.

Furthermore, floating point formats in intermediate layers provide better results for the same amount of bits than fixed point formats. Future research include determining what the optimal architecture should be to balance the LUT size and the number of operations for each inference.

\bibliography{cnn2,coding_theory,sensors,quant,markov,consensus,secure,synch,misc,stability,cml,algebraic_graph,graph_theory,control,optimization,adaptive,top_conjugacy,ckt_theory,math,number_theory,matrices,halftoning,image_processing,ai,communications,cad,sequences,computing,neural_nets}

\end{document}